\documentclass{article}
\usepackage{spconf,amsmath,amssymb,graphicx,multirow}

\usepackage{enumitem}
\usepackage{svg}

\usepackage{algorithm}
\usepackage{algpseudocode}
\usepackage{booktabs}
\usepackage{float}

\setlist{nosep, leftmargin=14pt}

\usepackage{mwe} 

\usepackage{eso-pic}
\newcommand{\ieeecopyrightnotice}{%
  \AddToShipoutPictureFG*{%
    \AtPageLowerLeft{%
      \hspace*{\dimexpr\oddsidemargin+1in\relax}%
      \raisebox{1cm}{%
        \parbox{0.95\textwidth}{\raggedright\footnotesize
          \copyright\ 2026 IEEE. Personal use of this material is permitted. Permission from IEEE must be obtained for all other uses, in any current or future media, including reprinting/republishing this material for advertising or promotional purposes, creating new collective works, for resale or redistribution to servers or lists, or reuse of any copyrighted component of this work in other works.}%
      }%
    }%
  }%
}

\title{Learning to Look Closer: A New Instance-wise Loss for Small Cerebral Lesion Segmentation}
\name{\begin{tabular}[t]{@{}c@{}}
  Luc Bouteille\textsuperscript{1} \qquad
  Alexander Jaus\textsuperscript{2} \\
  Jens Kleesiek\textsuperscript{1} \qquad
  Rainer Stiefelhagen\textsuperscript{2} \qquad
  Lukas Heine\textsuperscript{1} \qquad
  \end{tabular}
}

\address{\textsuperscript{1} Institute for AI in Medicine (IKIM), University Hospital Essen (A{\"o}R), Essen, Germany\\
  \textsuperscript{2} Institute for Anthropomatics and Robotics (IAR), Karlsruhe Institute of Technology, Karlsruhe, Germany
}
\begin{document}
\maketitle
\ieeecopyrightnotice
\begin{abstract}

Traditional loss functions in medical image segmentation, such as Dice, often under-segment small lesions because their small relative volume contributes negligibly to the overall loss. To address this, instance-wise loss functions and metrics have been proposed to evaluate segmentation quality on a per-lesion basis. We introduce CC-DiceCE, a loss function based on the CC-Metrics framework, and compare it with the existing blob loss. Both are benchmarked against a DiceCE baseline within the nnU-Net framework, which provides a robust and standardized setup. We find that CC-DiceCE loss increases detection (recall) with minimal to no degradation in segmentation performance, though with dataset-dependent trade-offs in precision. Furthermore, our multi-dataset study shows that CC-DiceCE generally outperforms blob loss.

\end{abstract}
\section{Introduction}
\label{sec:intro}

\begin{figure}[ht]
\centering
\begin{minipage}[b]{.48\linewidth}
  \centering
  \includegraphics[width=\linewidth]{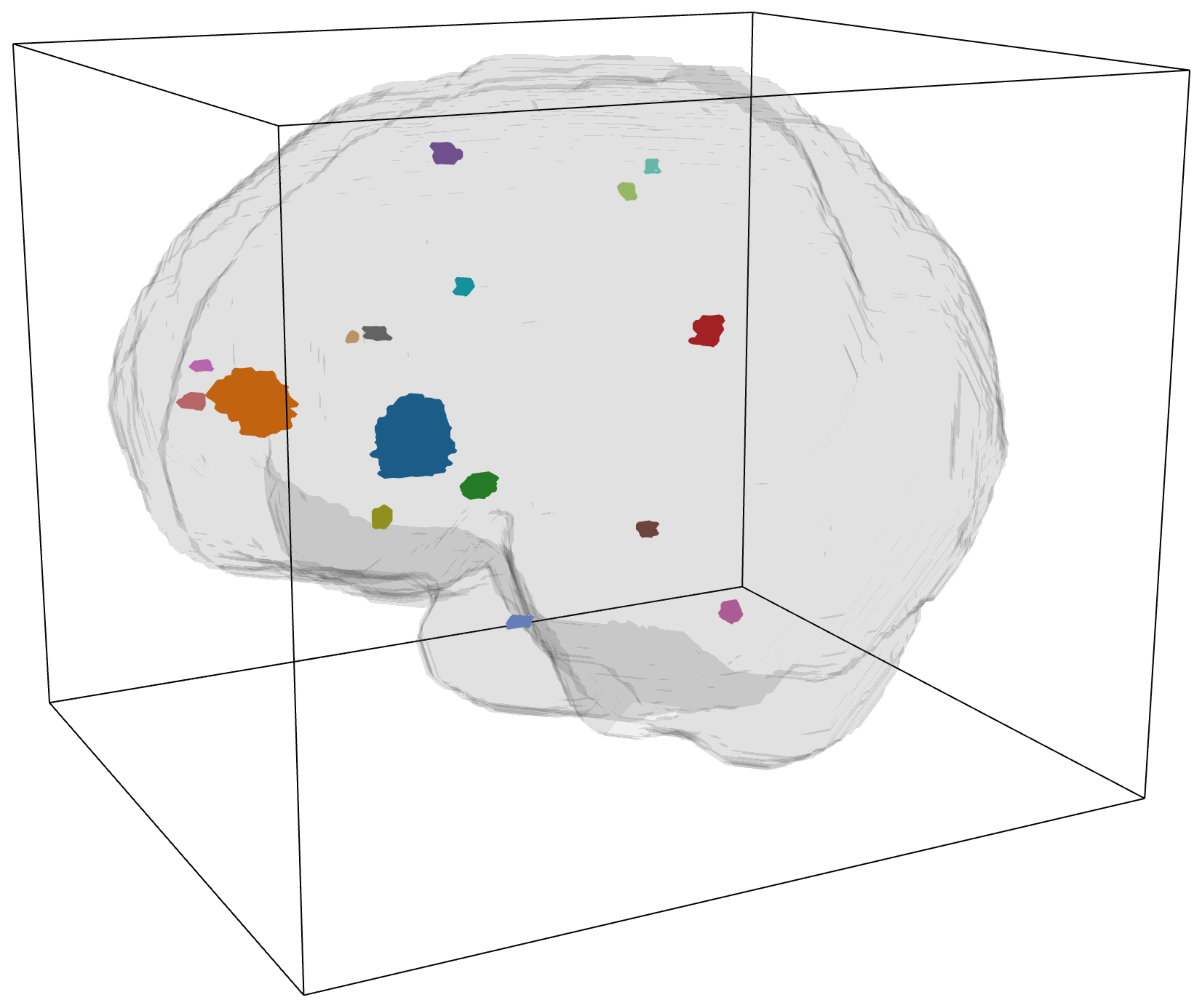}\\[2pt]
  {\small\shortstack[c]{Dice: 0.0\\
    CC-Dice: 0.0\\
    Blob-Dice: 0.0}}
\end{minipage}\hfill
\begin{minipage}[b]{.48\linewidth}
  \centering
  \includegraphics[width=\linewidth]{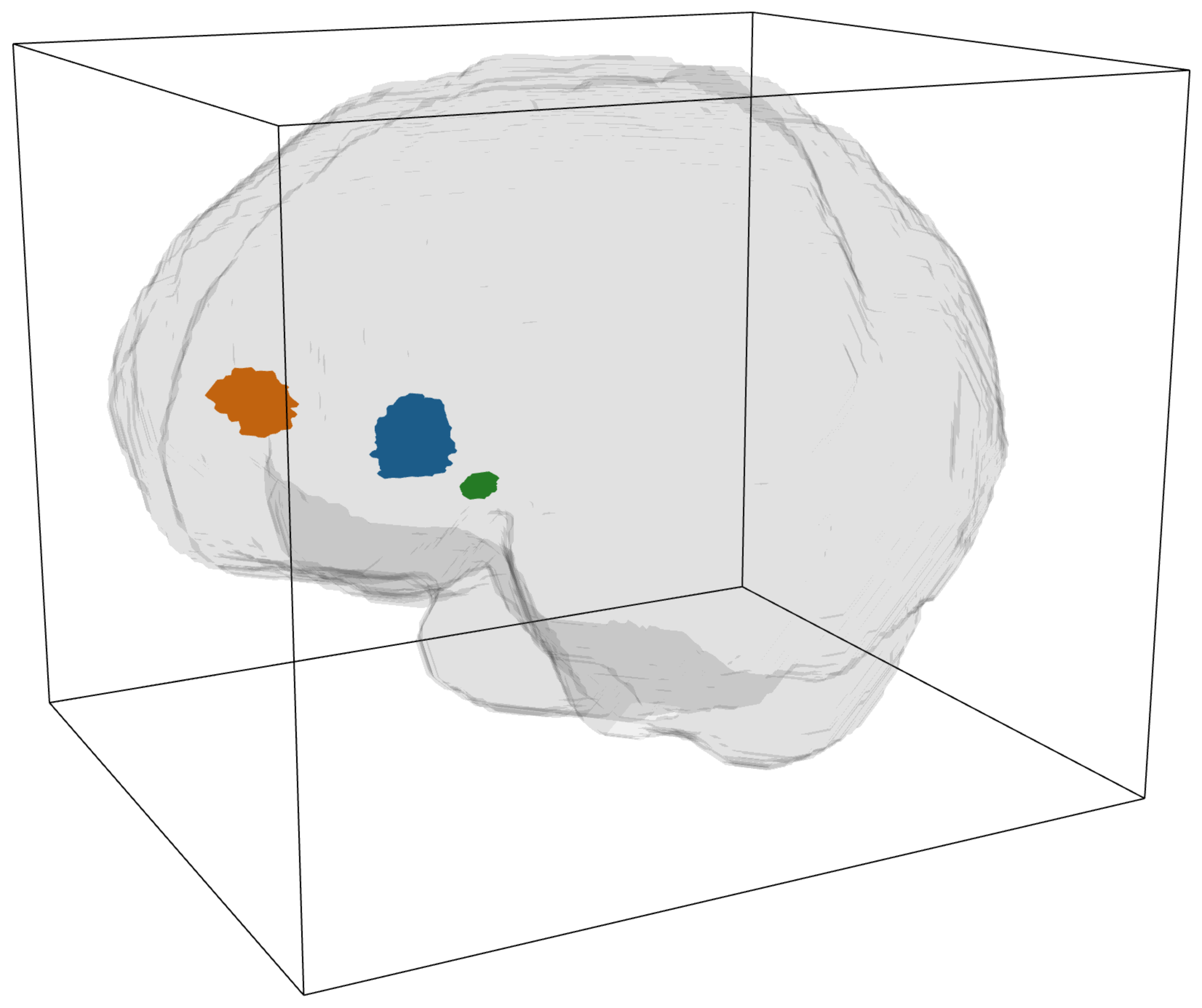}\\[2pt]
  {\small\shortstack[c]{Dice: 0.061\\
    CC-Dice: 0.8125\\
    Blob-Dice: 0.8125}}
\end{minipage}
\caption{Example from the Stanford BrainMetShare dataset \cite{sbm}. Left: perfect segmentation, each lesion shown in a different color. Right: imperfect prediction. Losses are listed beneath each rendering. The brain hull is schematic and not anatomically accurate.}
\label{fig:brain-hulls}
\end{figure}

Automated segmentation of small cerebral lesions in brain MRI enables scalable detection and quantification \cite{brats2023}. However, conventional voxel-overlap losses, such as the standard Dice with cross-entropy (DiceCE) used in nnU-Net \cite{nnunet}, overweight large structures. This can reduce small-lesion detection when lesion sizes vary widely \cite{blobloss, ccmetrics, instanceloss}, as small, clinically relevant instances contribute negligibly to the loss gradient \cite{brats2023, blobloss}. For example, in the Stanford BrainMetShare (SBM) dataset \cite{sbm}, brain metastasis volumes range from \(4.4\,\mathrm{mm}^3\) to \(779.4\,\mathrm{mm}^3\) (\(P_5\) vs. \(P_{95}\)). As illustrated in Fig. \ref{fig:brain-hulls}, this size discrepancy means that small instances have little effect on the Dice loss: the right-hand prediction correctly segments only 3 of 16 instances, yet the Dice loss remains close to zero.

\noindent To improve detection rates, instance-wise loss functions evaluate metrics on a per-component basis. Prior works include blob loss \cite{blobloss}, the Instance-wise and Center-of-Instance (ICI), False, and Proximity losses \cite{instanceloss, loss-instance-center}, all aiming to improve detection. However, comparison across these works is difficult because their experimental protocols differ substantially in architecture, dimensionality, datasets, and training pipeline. For example, the original ICI study used a 3D residual U-Net on one dataset, the later study introducing False and Proximity reported 2D SwinUNETR experiments on three datasets, and blob loss was evaluated in a separate 3D U-Net pipeline. Moreover, \cite{instanceloss} reported per-batch training times of 17.7--23.6\,s for the ICI, False, and Proximity losses, versus 0.66--0.77\,s for blob loss \cite{instanceloss}, making them impractical for our benchmark. As emphasized by \cite{nnunetrevisited}, rigorous and standardized evaluation is essential in medical image segmentation given dataset heterogeneity. We therefore perform a standardized multi-dataset evaluation of instance-aware losses within nnU-Net and additionally propose adapting CC-Metrics \cite{ccmetrics} as a loss function. Thus, our contributions are:
\begin{enumerate}
\item We investigate the potential of CC-Metrics as a loss function for small-lesion segmentation.
\item We provide a rigorous evaluation of instance-aware losses (CC-Metrics and blob loss) against a strong, standardized baseline (nnU-Net) across multiple heterogeneous datasets.
\end{enumerate}
\noindent The code for the experiments can be found at\\https://github.com/TIO-IKIM/Learning-to-Look-Closer.

\section{Related Work}
\label{sec:related-work}

\subsection{Losses}

Let $\mathbb{T}\subset\mathbb{Z}^3$ be the lattice, $K\subset\mathbb{T}$ the ground-truth foreground, $\mathcal{K}$ the set of its maximal 26-connected components, and $m:\mathcal{P}(\mathbb{T})\times\mathcal{P}(\mathbb{T})\to\mathbb{R}$ a base set metric (e.g., Dice). For $t\in\mathbb{T}$ and $A\subseteq\mathbb{T}$, define $d(t,A)=\min_{u\in A}\|t-u\|_2$. For each $C\in\mathcal{K}$, the Voronoi region is (ties broken arbitrarily)
\begin{equation}
R_C=\bigl\{\,t\in\mathbb{T}\;:\; d(t,C)<d(t,C'),\ 
\forall\,C'\in\mathcal{K}\setminus\{C\}\,\bigr\}.
\end{equation}
\noindent CC-Metrics computes the Voronoi region for every connected component (lesion) and then scores each region individually. This assigns the same weight to every lesion, regardless of size, in the final score.
\begin{equation}
\label{eq:cc_metrics}
V_m(P,K)=\frac{1}{|\mathcal{K}|}\sum_{C\in\mathcal{K}} m\bigl(P\cap R_C,\; C\bigr).
\end{equation}

\begin{figure}[htb]
\centering
\begin{minipage}[b]{.48\linewidth}
  \centering
  \includegraphics[width=\linewidth]{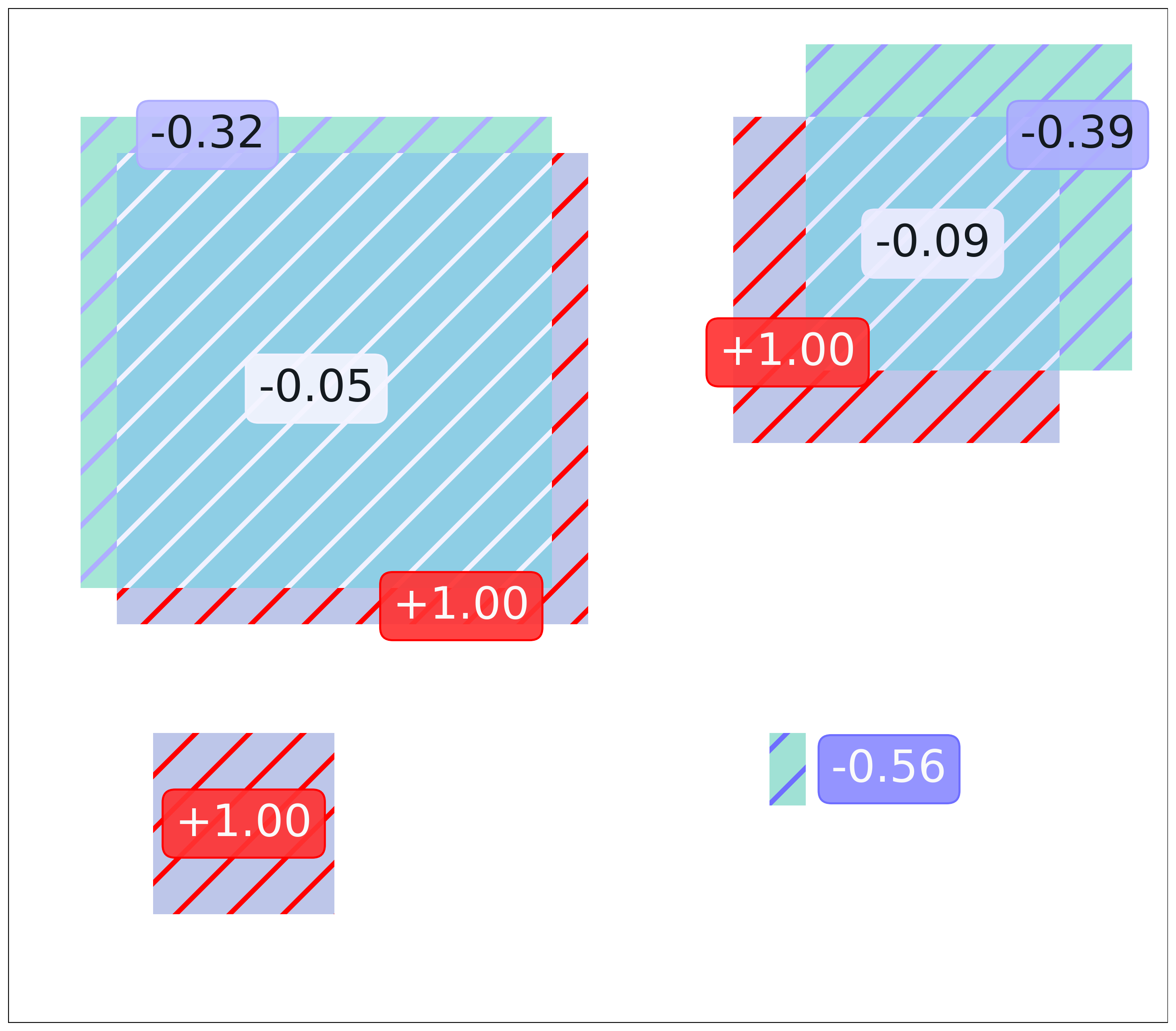}\\[2pt]
\end{minipage}\hfill
\begin{minipage}[b]{.48\linewidth}
  \centering
  \includegraphics[width=\linewidth]{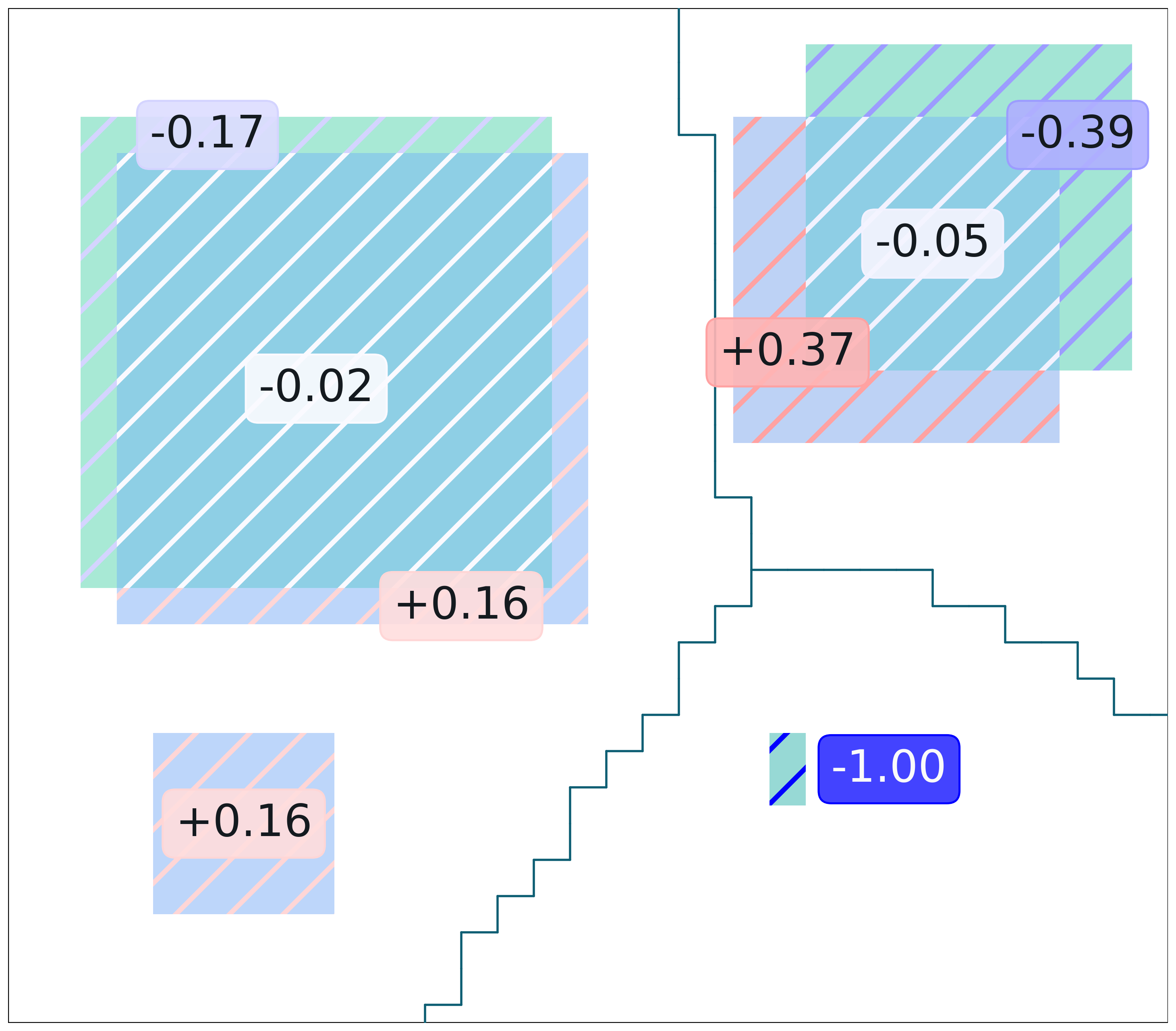}\\[2pt]
\end{minipage}
\caption{Per-voxel gradients with respect to logits for BlobDiceCE (left) and CC-DiceCE (right). Green rectangles denote ground truth, and light blue rectangles denote prediction. Shading color encodes gradient sign and darkness encodes magnitude. Values are normalized per panel. Voronoi regions are shown on the right. Normalized gradient values are shown in text boxes.}
\label{fig:gradient-illustration}
\end{figure}

\noindent In contrast, blob loss avoids geometric partitioning and instead isolates each component $C$ by masking. For each $C\in\mathcal{K}$, we zero out predictions on voxels belonging to other GT components (preserving false positives) and define the loss as the average over all components:
\begin{equation}
\label{eq:blob_loss}
B_m(P,K)=\frac{1}{|\mathcal{K}|}\sum_{C\in\mathcal{K}} m\bigl(P\setminus (K\setminus C),\; C\bigr).
\end{equation}
\noindent Fig. \ref{fig:gradient-illustration} contrasts the gradients of both loss functions. We observe two effects. First, in BlobDiceCE, false positives (dark red areas) have uniform gradients across component sizes and larger relative magnitudes than in CC-DiceCE. Second, in CC-DiceCE, gradients scale with Voronoi size, so smaller regions receive larger relative magnitudes. The small false negative in the bottom right therefore carries a larger gradient relative to the other regions under CC-DiceCE (\(-1.00\)), whereas in BlobDiceCE its effect is diluted (\(-0.56\)) because all false positives influence the computation for every component.

\subsection{Datasets}

We used five publicly available brain MRI datasets, deliberately selected to span a wide range of pathologies, lesion sizes, and lesion multiplicities to stress-test instance-aware objectives.
\noindent\textbf{Lacunes (LAC):} Lacunes are small ($\approx 3 \text{--} 15\,\text{mm}$), rounded subcortical cavities with CSF-like signal on MRI \cite{lacunesdef}. They represent the chronic sequelae of small deep infarcts and, less often, tiny hemorrhages in territories of penetrating arterioles. This dataset is from the ``Where is VALDO?'' \cite{wherevaldo} challenge and consists of 40 scans in T1, T2, and FLAIR modalities from two sources.

\noindent\textbf{Cerebral Microbleeds (CMB):} Cerebral microbleeds are tiny (\( \leq 10\,\mathrm{mm} \)) foci of chronic blood products within the brain parenchyma that appear as small, well-circumscribed signal-loss dots on T2*/SWI. The underlying sources for LAC and CMB are the same. There are 72 scans in total from three different cohorts. 

\noindent\textbf{Stanford BrainMetShare (SBM):} Brain metastases are a frequent neurologic complication of cancer, typically originating from lung tumors, breast tumors, and malignant melanoma \cite{sbm}. The dataset consists of 105 scans with MRI inputs including pre- and post-gadolinium T1-weighted 3D fast spin echo (CUBE), post-gadolinium T1-weighted 3D axial IR-prepped FSPGR (BRAVO), and 3D CUBE FLAIR.

\noindent\textbf{White Matter Hyperintensities Challenge (WMH):} White matter hyperintensities are focal or confluent areas of increased signal on T2-weighted/FLAIR MRI and represent a hallmark of cerebral small vessel disease. This dataset consists of 60 scans in T1 and FLAIR modalities. It originates from the White Matter Hyperintensities Challenge \cite{wmhchallenge}.

\noindent\textbf{BraTS:} Gliomas are the most common primary malignant tumors of the central nervous system
and exhibit substantial clinical and radiographic heterogeneity. The BraTS 2020 challenge \cite{brats1, brats2, brats3, brats4} comprises brain MRI scans with expert labels for gliomas. The included modalities are T1, T1Gd, T2, and T2-FLAIR. The dataset contains 369 training cases.\\\\
For instance-aware losses, the distribution of connected components (CCs), i.e., 3D foreground objects in the binarized ground truth, is a key driver of optimization behavior. Tab. \ref{tab:dataset-stats} summarizes per-scan connected component (CC) counts and object sizes across datasets. We report the median number of CCs per case with interquartile range (``CC \( P_{50}\,[P_{25}, P_{75}] \)'') and the mean CC volume with its standard deviation. Volumes are computed in \( \mathrm{mm}^3 \) using the original voxel spacing of each dataset. CCs are obtained via standard 26-connectivity connected-component labeling on the binarized masks. Across all datasets, two properties stand out: First, they exhibit heavy-tailed object-size distributions. The standard deviation of CC volumes is large relative to the mean (see Tab. \ref{tab:dataset-stats}), indicating that a few large components coexist with many small ones. Second, lesion multiplicity varies by task. LAC and CMB typically contain zero to a few CCs per scan. SBM contains a moderate number. WMH contains dozens. BraTS contains few but very large CCs.

\begin{table}[ht]
\centering
\small
\begin{tabular}{lll}
\toprule
 & \multicolumn{1}{c}{CC \( P_{50}\,[P_{25}, P_{75}] \)} &
 \multicolumn{1}{c}{Mean volume $\pm$ std. [$\mathrm{mm}^3$]} \\
\midrule
LAC   & 1 [0, 2.25]         & 59.7 $\pm$ 85.4 \\
CMB   & 1 [0, 2]            & 18.7 $\pm$ 26 \\
SBM   & 6 [2, 14]           & 240.5 $\pm$ 1355.2 \\
WMH   & 53.5 [36.5, 81.25]  & 287.1 $\pm$ 2074.7 \\
BraTS & 3 [2, 7]            & 17997.4 $\pm$ 45516.9 \\
\bottomrule
\end{tabular}
\caption{Per-scan connected component counts and component volumes across datasets.} 
\label{tab:dataset-stats}
\end{table}

\section{Experimental Setup}
\label{sec:experiments}

\subsection{Training}

We evaluated all loss functions within the nnU-Net framework. We retained all default nnU-Net parameters, with one exception: we used the non-smooth variant of Dice loss with \( \epsilon = 0 \), as we observed training instability on the CMB dataset with the default smooth Dice. The base metric \( m \) for the instance-wise functions was DiceCE. As recommended in \cite{blobloss,loss-instance-center}, we combined the instance-wise losses (Blob and CC) with the global DiceCE loss in a \(1{:}1\) ratio, resulting in the \textit{BlobDiceCE} and \textit{CC-DiceCE} objectives. All experiments used five-fold cross-validation.

\subsection{Datasets}

We applied small, dataset-specific adjustments to harmonize evaluation with the binary, instance-aware objectives. LAC contains two raters; their masks were logical OR-combined so that any
voxel marked by either rater was considered a lacune. BraTS defines three labels; we binarized them by collapsing all tumor labels into one positive class. WMH includes two labels; we omitted the auxiliary label so that training and evaluation remained binary. For all datasets, we used all provided modalities and only the training sets, since the test sets are not available for every dataset.

\subsection{Evaluation}

We measured three metrics: Dice, CC-Dice, and instance-wise F1. For F1, a true positive was defined when a ground-truth and a predicted connected component had an overlap of at least one voxel using one-to-one matching. For each fold, we computed these metrics on the corresponding validation split. We report the mean and standard deviation across the five folds. Furthermore, we report recall per volume quartile using the same overlap criterion, as shown in Tab. \ref{tab:combined-metrics}.

\section{Results}
\label{sec:results}

\begin{table*}[t] \centering
\caption{Combined per-dataset metrics for three loss functions. Left: Overall metrics (mean $\pm$ std.). Right: Recall by lesion volume quartile.}
\label{tab:combined-metrics}
\setlength{\tabcolsep}{3pt} \scriptsize \begin{tabular}{l l c c c | l c c c}
\toprule
\multirow{2}{*}{\textbf{Dataset}} & \multirow{2}{*}{\textbf{Metric}} & \multicolumn{3}{c}{\textbf{Results per Loss Function}} & \multirow{2}{*}{\textbf{Quartile}} & \multicolumn{3}{c}{\textbf{Recall by Volume Quartile}} \\
\cmidrule(lr){3-5} \cmidrule(lr){7-9} & & \textbf{DiceCE} & \textbf{BlobDiceCE} & \textbf{CC-DiceCE} & & \textbf{DiceCE} & \textbf{BlobDiceCE} & \textbf{CC-DiceCE} \\
\midrule

\multirow{5}{*}{LAC} 
& Dice & $\mathbf{0.2770} \pm\,0.1796$ & $0.2504 \pm\,0.1779$ & $\underline{0.2657} \pm\,0.1622$ 
& 0-25\% & $\underline{0.0000}$ & $\underline{0.0000}$ & $\mathbf{0.0588}$ \\
& CC-Dice & $\underline{0.1942} \pm\,0.1263$ & $0.1871 \pm\,0.1441$ & $\mathbf{0.2010} \pm\,0.1302$ 
& 25-50\% & $\mathbf{0.3529}$ & $\underline{0.2353}$ & $\mathbf{0.3529}$ \\
& Precision & $\underline{0.3747} \pm\,0.1946$ & $0.3364 \pm\,0.1835$ & $\mathbf{0.4291} \pm\,0.1311$ 
& 50-75\% & $\underline{0.7059}$ & $\mathbf{0.7647}$ & $0.6471$ \\
& Recall & $\underline{0.3211} \pm\,0.1819$ & $0.3031 \pm\,0.1912$ & $\mathbf{0.3698} \pm\,0.1163$ 
& 75-100\% & $\underline{0.5882}$ & $\underline{0.5882}$ & $\mathbf{0.6471}$ \\
& F1 & $\underline{0.3334} \pm\,0.1787$ & $0.2993 \pm\,0.1693$ & $\mathbf{0.3661} \pm\,0.0856$ 
& & & & \\
\midrule

\multirow{5}{*}{CMB} 
& Dice & $0.3951 \pm\,0.0519$ & $\underline{0.4072} \pm\,0.0752$ & $\mathbf{0.4245} \pm\,0.0597$ 
& 0-25\% & $\mathbf{0.2623}$ & $\underline{0.2459}$ & $\underline{0.2459}$ \\
& CC-Dice & $0.3655 \pm\,0.0673$ & $\underline{0.3805} \pm\,0.0925$ & $\mathbf{0.3920} \pm\,0.0650$ 
& 25-50\% & $0.2787$ & $\underline{0.3115}$ & $\mathbf{0.3279}$ \\
& Precision & $0.5335 \pm\,0.0761$ & $\underline{0.5614} \pm\,0.0944$ & $\mathbf{0.6014} \pm\,0.0345$ 
& 50-75\% & $\mathbf{0.6491}$ & $\underline{0.6140}$ & $\mathbf{0.6491}$ \\
& Recall & $\underline{0.5417} \pm\,0.1331$ & $0.5386 \pm\,0.1322$ & $\mathbf{0.5780} \pm\,0.1188$ 
& 75-100\% & $\underline{0.7193}$ & $\underline{0.7193}$ & $\mathbf{0.7719}$ \\
& F1 & $0.5101 \pm\,0.0830$ & $\underline{0.5254} \pm\,0.0951$ & $\mathbf{0.5616} \pm\,0.0682$ 
& & & & \\
\midrule

\multirow{5}{*}{SBM} 
& Dice & $\underline{0.6749} \pm\,0.0214$ & $0.6458 \pm\,0.0173$ & $\mathbf{0.6753} \pm\,0.0160$ 
& 0-25\% & $\underline{0.3848}$ & $0.3701$ & $\mathbf{0.5074}$ \\
& CC-Dice & $\underline{0.5343} \pm\,0.0244$ & $0.5101 \pm\,0.0260$ & $\mathbf{0.5462} \pm\,0.0238$ 
& 25-50\% & $\underline{0.6947}$ & $0.6218$ & $\mathbf{0.7563}$ \\
& Precision & $\underline{0.8452} \pm\,0.0325$ & $\mathbf{0.8463} \pm\,0.0611$ & $0.8431 \pm\,0.0233$ 
& 50-75\% & $\underline{0.8182}$ & $0.7914$ & $\mathbf{0.8316}$ \\
& Recall & $\underline{0.7994} \pm\,0.0266$ & $0.7780 \pm\,0.0222$ & $\mathbf{0.8115} \pm\,0.0214$ 
& 75-100\% & $\underline{0.9180}$ & $0.9101$ & $\mathbf{0.9206}$ \\
& F1 & $\mathbf{0.8011} \pm\,0.0102$ & $0.7884 \pm\,0.0383$ & $\underline{0.7999} \pm\,0.0098$ 
& & & & \\
\midrule

\multirow{5}{*}{WMH} 
& Dice & $\mathbf{0.7711} \pm\,0.0376$ & $0.7190 \pm\,0.0471$ & $\underline{0.7705} \pm\,0.0370$ 
& 0-25\% & $\mathbf{0.4203}$ & $0.3696$ & $\underline{0.4095}$ \\
& CC-Dice & $\mathbf{0.4624} \pm\,0.0596$ & $0.4133 \pm\,0.0653$ & $\underline{0.4458} \pm\,0.0534$ 
& 25-50\% & $\mathbf{0.6104}$ & $0.5723$ & $\underline{0.5919}$ \\
& Precision & $\mathbf{0.7563} \pm\,0.0233$ & $0.7124 \pm\,0.0340$ & $\underline{0.7492} \pm\,0.0303$ 
& 50-75\% & $\underline{0.7732}$ & $0.7504$ & $\mathbf{0.7760}$ \\
& Recall & $\mathbf{0.6564} \pm\,0.0514$ & $0.6339 \pm\,0.0552$ & $\underline{0.6432} \pm\,0.0439$ 
& 75-100\% & $\underline{0.9110}$ & $\mathbf{0.9275}$ & $0.9099$ \\
& F1 & $\mathbf{0.6956} \pm\,0.0356$ & $0.6619 \pm\,0.0440$ & $\underline{0.6847} \pm\,0.0310$ 
& & & & \\
\midrule

\multirow{5}{*}{BraTS} 
& Dice & $\underline{0.9171} \pm\,0.0048$ & $0.9168 \pm\,0.0043$ & $\mathbf{0.9174} \pm\,0.0025$ 
& 0-25\% & $0.0057$ & $\underline{0.0071}$ & $\mathbf{0.0767}$ \\
& CC-Dice & $\underline{0.4179} \pm\,0.0119$ & $0.4160 \pm\,0.0137$ & $\mathbf{0.4443} \pm\,0.0153$ 
& 25-50\% & $0.0111$ & $\underline{0.0250}$ & $\mathbf{0.1333}$ \\
& Precision & $\mathbf{0.7708} \pm\,0.0233$ & $\underline{0.7159} \pm\,0.0618$ & $0.0517 \pm\,0.0093$ 
& 50-75\% & $0.0407$ & $\underline{0.0471}$ & $\mathbf{0.2334}$ \\
& Recall & $\underline{0.4513} \pm\,0.0141$ & $0.4510 \pm\,0.0112$ & $\mathbf{0.5286} \pm\,0.0230$ 
& 75-100\% & $0.7961$ & $\underline{0.8059}$ & $\mathbf{0.8549}$ \\
& F1 & $\mathbf{0.4933} \pm\,0.0122$ & $\underline{0.4789} \pm\,0.0253$ & $0.0841 \pm\,0.0086$ 
& & & & \\
\bottomrule
\end{tabular}
\end{table*}

As summarized in Tab. \ref{tab:combined-metrics}, replacing the baseline DiceCE with CC-DiceCE maintained the global Dice score within the typical 5-fold variation for each cohort, while improving CC-Dice and recall in most of them.\\On LAC and CMB, CC-DiceCE increased lesion-wise performance (higher CC-Dice and F1) with a trade-off in global Dice on LAC and a consistent gain on CMB. On SBM, Dice was essentially unchanged, while CC-Dice and recall improved; precision and F1 were within the baseline's typical variation. On WMH, Dice remained stable, but instance-aware metrics were lower. On BraTS, CC-Dice increased, but precision dropped notably, leading to a lower instance-wise F1 despite the increased recall.\\Compared to BlobDiceCE, CC-DiceCE generally matched or exceeded its instance-aware metrics while avoiding the Dice score reductions observed with BlobDiceCE on LAC, SBM, and WMH.\\The recall improvements from CC-DiceCE were not confined to small lesions. As shown in the quartile analysis in Tab. \ref{tab:combined-metrics}, recall gains were distributed similarly across all lesion size quartiles.

\section{Discussion}
\label{sec:discussion}

We found that CC-DiceCE improved detection rates (recall) while the change in segmentation performance (Dice) was minimal (at worst $-0.011$) across all five datasets. We also observed increases in CC-Dice in four of five datasets, with a small decrease only on WMH. We hypothesize that the inclusion of the global DiceCE loss term helped maintain the overall segmentation quality, which is consistent with findings in \cite{blobloss}. WMH stood out as the only dataset where CC-DiceCE showed no improvement in any metric compared with DiceCE. There were indications that the default nnU-Net configuration overfit on this dataset. With a shorter schedule of 150 epochs (instead of the default 1000), we obtained higher Dice for DiceCE ($0.799$) and CC-DiceCE ($0.792$) and higher CC-Dice ($0.496$ and $0.535$). This suggests that under the default schedule, overfitting may have obscured potential CC-DiceCE gains. We observed similar overfitting with BlobDiceCE, which yielded a Dice of $0.770$ at 150 epochs.\\\\The theoretical properties of CC-Metrics offer an explanation for the increased detection rates. Missing a ground-truth component (a false negative) sets its $1/N$ contribution to the loss to the worst possible value, a significant penalty. This mechanism encourages higher recall, which we observed in practice on four of the five datasets. Conversely, a false positive typically has a smaller effect, as it only impacts the score of the single Voronoi region it occupies. This design did not lead to a consistent drop in precision. The largest decrease occurred on BraTS, likely due to its label structure, where gliomas often form a single large component with small annotation artifacts nearby. Under CC-Metrics, missing these small components is strongly penalized, whereas adding small peripheral false positives only slightly lowers the score of the affected Voronoi region. This effect is less pronounced with BlobDiceCE, which, as discussed in Sec. \ref{sec:related-work}, more strongly penalizes false positives. We initially hypothesized that CC-DiceCE would primarily aid the detection of small lesions. However, our results (Tab. \ref{tab:combined-metrics}) showed that recall improvements were distributed across all size quartiles, not just the smallest. This suggests that CC-DiceCE acts as a general instance regularizer, not just a small-object detector.\\Overall, CC-DiceCE tended to outperform blob loss across all five datasets. It showed improvements in 22 out of 25 metric-dataset combinations. Blob loss also tended to show a larger decrease in segmentation performance (Dice) compared with CC-DiceCE.

\section{Conclusion}
\label{sec:conclusion}

We studied instance-aware objectives for small cerebral lesion segmentation within a strong and standardized nnU-Net setup across five heterogeneous MRI cohorts. Replacing conventional DiceCE with CC-DiceCE consistently improved instance-aware detection (higher recall and CC-Dice) in four of five datasets while having negligible effect on global Dice. In contrast, BlobDiceCE yielded mixed results and overall was outperformed by CC-DiceCE on most datasets. We hypothesize that CC-DiceCE amplifies the penalty of missed lesions, which encourages more aggressive detection and therefore higher recall, and can increase false positives, especially in tasks dominated by a few large components (e.g., BraTS and WMH). In practical clinical settings, an elevated false positive rate may be the lesser of two evils, ensuring that no clinically relevant lesions are missed and allowing radiologists to verify detections rather than risk overlooking subtle but critical findings. We also found that CC-DiceCE improved detection across lesion sizes rather than only in small lesions.\\In general, the results suggest that CC-DiceCE provides a simple and effective objective to improve lesion performance while preserving global overlap.\\Further work should expand the datasets to other modalities and pathologies to investigate CC-DiceCE's potential in anatomical regions outside the brain.

\section{Compliance with Ethical Standards}
\label{sec:ethics}

This research study was conducted retrospectively using human subject data made available in open access \cite{wherevaldo,wmhchallenge,sbm, brats1, brats2, brats3, brats4}. Ethical approval was not required
as confirmed by the license attached with the open access
data.
\pagebreak
\bibliographystyle{IEEEbib}
\bibliography{refs}

\end{document}